\documentclass[10pt,twocolumn,letterpaper]{article}

\usepackage{cvpr}
\usepackage{times}
\usepackage{epsfig}
\usepackage{graphicx}
\usepackage{amsmath}
\usepackage{amssymb}

\usepackage{booktabs} 
\usepackage{multirow}
\usepackage{pifont}% http://ctan.org/pkg/pifont

\newcommand{\ourmethod}{DEGAS}

\usepackage[pagebackref=true,breaklinks=true,letterpaper=true,colorlinks,bookmarks=false]{hyperref}

\cvprfinalcopy % *** Uncomment this line for the final submission

\def\cvprPaperID{8199} % *** Enter the CVPR Paper ID here

\ifcvprfinal\fi
\begin{document}

%%%%%%%%% TITLE
\title{\ourmethod{}: Differentiable Efficient Generator Search }

\author{Sivan Doveh and 
Raja Giryes
\\ \\
Tel Aviv University, Israel}
% \and

\maketitle
\let\thefootnote\relax\footnotetext{\\Corresponding author: Sivan Doveh \\ \texttt{sivandoveh@mail.tau.ac.il}}

\begin{abstract}
Network architecture search (NAS) achieves state-of-the-art results in various tasks such as classification and semantic segmentation. Recently, a reinforcement learning-based approach has been proposed for Generative Adversarial Networks (GANs) search.
In this work, we propose an alternative strategy for GAN search by using a method called DEGAS (Differentiable Efficient GenerAtor Search), which focuses on efficiently finding the generator in the GAN.  
Our search algorithm is inspired by the differential architecture search strategy and the Global Latent Optimization (GLO) procedure. This leads to both an efficient and stable GAN search. 
After the generator architecture is found, it can be plugged into any existing framework for GAN training. For CTGAN, which we use in this work, the new model outperforms the original inception score results by 0.25 for CIFAR-10 and 0.77 for STL. It also gets better results than the RL based GAN search methods in shorter search time.
\end{abstract}

\section{Introduction}
\label{Introduction}
Generative Adversarial Networks (GANs) \cite{gan} have become a very successful framework for image generation. This scheme includes a generator that creates images and a discriminator that tries to discriminate between real and synthesized images. Both are trained together using a min-max based optimization.  

GANs are difficult to train, due to problems such as mode collapse, non-convergence to the Nash equilibrium and vanishing gradients. Overcoming these problems is the focus of many recent works. Notable among them is the Wasserstein GAN (WGAN) \cite{wgan}, which suggests replacing the KL-divergence loss with the earth mover's distance (Wasserstein) loss for solving the diminishing gradient problem, and is the basis for many other works \cite{wgan-gp,ctagn,cagan}. Another approach, GLO \cite{glo}, changes the traditional adversarial framework, by removing the discriminator and using a reconstruction loss of the input instead, which avoids many of the instabilities that appear in joint generator-discriminator training \cite{Kurach19Large, IS}. %A similar strategy has been also proposed in \cite{Hoshen18NAM}. 

The architectures that are used for the generator in the GAN works are usually inspired by DC-GAN \cite{dcgan} and ResNet \cite{resnet}. However, all of them are manually designed. 
In recent years, neural architecture search (NAS) based approaches have successfully found models that outperform the state-of-the-art in various tasks \cite{huang2018gpipe}. The first line of works used reinforcement learning \cite{zophNasRL} and genetic algorithms \cite{Real18Regularized} to generate a target classifier for a certain task. Yet, these approaches require a large number of computational resources, thus, making them computationally demanding. The second line of works manages to reduce the computational cost significantly to only a few days on a single GPU. Two notable works among them are Efficient NAS \cite{ENAS} and Differentiable Architecture Search (DARTS) \cite{liu19darts}. 

{\bf Contribution.} 
In this work, we aim at finding a generator model using an efficient neural architecture search. 
% The operations within the generator, which include components such as up-sample operations, are different from a common convolution neural network for classification. The length of the generator is also much smaller. Thus, we can search all the layers in the network, unlike most of the currently exiting NAS methods (e.g., DARTS) that assume a repeating structure that builds the network and search only for it.
Similar to DARTS, we use the concept of learning the weights of connections between feature maps and then performing pruning to get the final network structure. 

While we use the DARTS MixedOp to connect between feature maps and the concept of connection based differentiable search,  
we have four major differences from it:

a. {\bf Global search.} DARTS uses the cell method, i.e., learning a repeating structure in the network and then concatenating the learned repeated structures to create the whole large network. We do not use the cell method: Although many generator models are built from several repeating operations, we exploit the relatively short length of the generator network and the efficiency of the search algorithm to make the search more flexible in finding new structures. Thus, we search for the whole network altogether.

b. {\bf Operation types.} Instead of operations that reduce the size of the data that are part of the DARTS search space, we use up-sample operations that increase the size of their input and are inspired by the DC-GAN and ResNet models. 

c. {\bf Operations' connections.} DARTS concatenate the outputs of cells to form the input of the next cell. To allow having DC-GAN and ResNet type networks, we eliminate the concatenation and forward only operations that are selected in the search.  This is crucial to get the results we present in our work. Yet, it requires changing the pruning phase that is used in DARTS. More details on this issue and the changes it requires, appear in Section~\ref{Method}.

d. {\bf The objective function.} While DARTS uses classification losses in its training, we use a different objective. % However, we do not search for cells and concatenate them. Instead, we search for the whole network together. We also change the search operations to those that are more suitable for the generator architecture.
% Another important difference is the loss function, which is very different in GAN training compared to classification (which is the main problem considered in DARTS and other NAS techniques). 
Since their introduction, GANs have been very popular for image generation. Yet, their training is considered to be a difficult task. The GAN framework is composed of two components: (i) a generator that tries to generate images that look real, and (ii) a discriminator that discriminates between generated and real images. Learning both architectures simultaneously can turn into a very hard problem. In this work, we relax this problem by searching only for generator architecture. Moreover, to overcome the sensitivity of the GAN training, we search for the generator model using the GLO training protocol, i.e., search for the generator using a reconstruction loss. This gives us a very important advantage: The generator search is decoupled from the discriminator and thus we do not need to deal in the search with an untrained discriminator that will have a negative influence on the learning of the generator.

While we use GLO for the search, we do not use it for the final training of the generator since the visual results produced by GANs that have a discriminator are more favorable. Thus, to be on par with other existing GANs, we train the found generator architecture using an existing framework for GAN training. Specifically, we use the CTGAN protocol \cite{ctagn} and show that our automatically designed generator improves the performance in this framework compared to the original manually designed GAN used with it both in terms of FID \cite{fid} and inception scores \cite{IS}.
By this, we show that a generator search that is based on image reconstruction is a valid strategy for finding generators that produce good images.

On all the datasets used in this work, the search took only a couple of days on a single GPU, %achieving results that are on par with other methods that were configured manually.
achieving better results than \cite{agan} and \cite{Autogan} that use reinforcement learning. Moreover, when we plug our generator to an existing GAN procedure, we also improve the results compared to the vanilla generator (in CTGAN).
This further demonstrates the advantage of the proposed approach for the generator search. 

\section{Background}
\label{Background}
{\bf Generative Adversarial Networks.}
The Generative Adversarial Networks (GANs) models \cite{gan} include a generator that synthesizes new data given a latent representation and
a discriminator, whose role is to discriminate between real data and synthesized data. The generator
and the discriminator are trained together using an adversarial game between the generator that tries to ‘fool’ the discriminator, which on its side tries to discriminate between true and synthesized examples. This is implemented by a min-max optimization problem whose goal is to reach the optimal point, which is the Nash equilibrium.

The common practice shows that training GANs is a difficult task. Frequent problems in their training include mode collapse, diminishing gradients and non-convergence to the Nash equilibrium. These hinders can cause significant difficulty in performing an automatic search for generator architecture. Much attention has been given recently to find a solution to these deficiencies. 

The Wasserstein GAN (WGAN) \cite{wgan} aims at reducing the imbalance between the discriminator and the generator in the training, in which the discriminator outperforms the generator. When this happens, the generator gradients according to the Jensen or KL divergence are zero, and thus it does not learn anything. To solve this, the Wasserstein loss function (known also as the earth mover's distance) is used. For using it, the discriminator function needs to be a differentiable 1-Lipchitz function. 

In the original WGAN paper, the 1-Lipchitz property has been enforced by clipping the weights of the network using a hyperparameter, which leads to a network that is very sensitive to the tuning of this hyperparameter. To fix this, the use of gradient penalty (GP) has been proposed \cite{wgan-gp}, under the assumption that the gradient of a 1-Lipchitz function needs to be one almost everywhere. This condition is added as a regularizer and leads to the WGAN-GP technique. 

Another notable work is CAGAN \cite{cagan}, which introduced the adversarial consistency loss. They used the training process of WGAN-GP with a large number of critics (different discriminators). Each critic was created by using a dropout on the hidden layers of the discriminator. They presented a consistency loss that reduced the redundancy between these critics.

More recently, CTGAN has been proposed \cite{ctagn}, which adds regularization on the discriminator loss function. The loss takes two random perturbations of the inputs to the discriminator and adds a penalty on the distance between their two outputs. In this work, we use the CTGAN training framework and discriminator for testing the generator architecture we have found.

\begin{figure}  
    \centering
        \includegraphics[width=\linewidth]{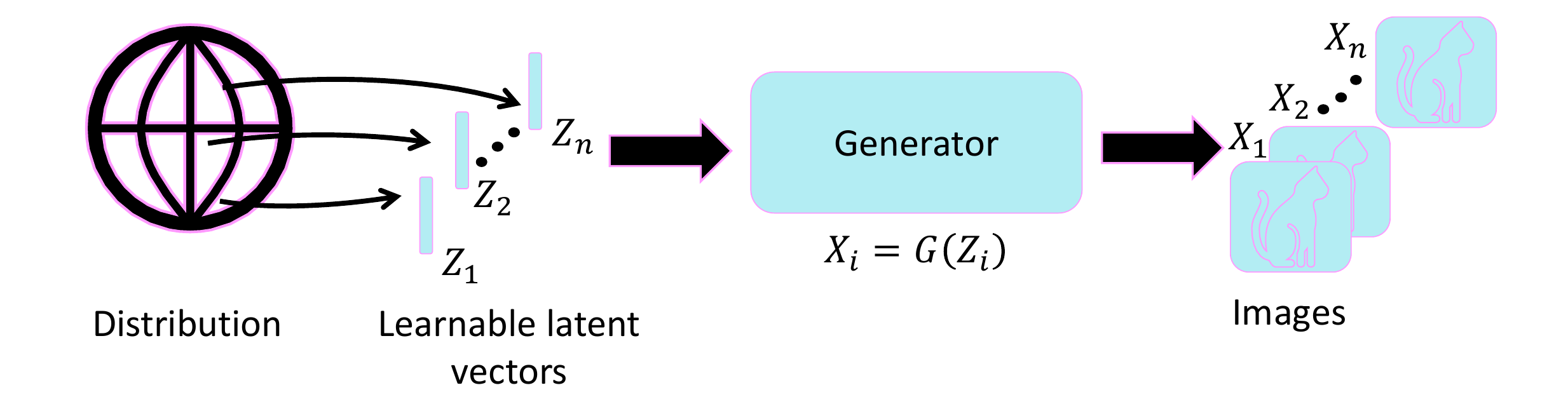}
        \caption{GLO method: each randomly sampled Z is matched to specific train image (X) and Z, G(z)=X are being learned}
        \label{fig:GLO}
    \end{figure}

The Generative Latent Optimization (GLO) framework \cite{glo} takes an alternative route for bypassing the challenge of training both the generator and the discriminator networks. It replaces the adversarial training that requires having a discriminator, with another optimization strategy that only trains a generator. This approach pairs to each image in the training set a latent representation and then optimizes the generator to decode each latent vector to its corresponding image (see Fig.~\ref{fig:GLO}). This strategy allows training generative models very easily without the need for a discriminator training. 
Another related approach is NAM \cite{Hoshen18NAM}, which is an unsupervised method for mapping without adversarial training.
In this method, a pre-trained generative model aligns each source image with a synthesized image from the target domain. This optimization is done together with the optimization of the domain mapping function.

Note that the overall performances of the GLO strategy in terms of image quality are not as good as training a generator
with a discriminator. However, the simplicity of this method is perfect for generator architecture search and therefore we use it in our work in the search phase. 

{\bf Architecture search techniques.} Several methods have been proposed for optimizing the parameters of neural networks and for finding new architectures. 
In \cite{Andrychowicz16Learning,Chen17Learning}, it has been shown that one may unfold a recurrent neural network to approximate some given functionals in a better way than other manually designed approaches. These strategies may allow better tuning of the hyper-parameters of a given network compared to Bayesian approaches.
Yet, they are not designed searching for new neural architectures. 

Zoph \emph{et al.} \cite{zophNasRL} used a reinforcement learning-based approach for neural architecture search (NAS). A recurrent network was used to generate the model description of a target neural network for a certain task. They showed improvement in their resulted architecture (NASnet) comparing to existing hand-crafted network models at its time. They were outperformed by AmoebaNet \cite{Real18Regularized}.  The work in \cite{Real17Large,Real18Regularized} introduces a different technique for finding neural architectures. Given a task, they use an evolutionary (genetic) algorithm to find a neural structure.  

Another approach for improving accuracy was proposed in \cite{AutoAug}. New data augmentation techniques for neural network training have been sought for using reinforcement learning. While this search process is computationally demanding, it has been demonstrated that the found augmentation techniques, even when produced by a relatively small dataset, are transferable across different problems (e.g., augmentation developed for CIFAR-10 were useful for ImageNet). 

All these methods require very large computational resources (some needs thousands of GPU days!),
therefore although they achieve state-of-the-art performances, in practice it is unfeasible to use them.

Recent works have managed to overcome the high computational cost, without reducing the neural architectures performances significantly. Notable among them are the Efficient NAS (ENAS) \cite{ENAS} and the Differentiable Architecture Search (DARTS) \cite{liu19darts}. Both of these work manage to search neural architectures in only a few GPU days.

ENAS is a reinforcement learning-based method. A sub-graph (child model) is searched in a large graph and to reduce the search time, the child models share weights between the same operations.

DARTS consists of two phases. Its network is built from a concatenation of cells, which are learned during the first phase. Then, in the second phase, some of the graph connections and operations are reduced according to a pruning algorithm. Each operation has a learned weight multiplier, which indicates connection importance. They are learned as continues variables during the first phase. After the second phase, the remaining weights after the pruning are the ones with the highest weight multiplier magnitude.

DARTS uses a harsh pruning at the end, which makes the found architecture sub-optimal. This problem may be addressed by performing gradual pruning \cite{ASAP} or layer-wise search \cite{PD}, which leads to better results in less time.

% \citet{proxy} addresses the high GPU memory problem of direct architecture search on large-scale tasks.
% While most algorithms search on proxy tasks, they search for the target architecture by training with a gradient-based method and force only one path in the network to be active at run-time.

A one-shot low memory-efficient solution has been suggested in \cite{proxy}. It uses a learned binary mask to select only a single path of the network and load it on the GPU. Their strategy searches global architecture and not cells.  

NAS methods have been in used also for other problems than classification. These include semantic image segmentation \cite{deeplab}, medical image segmentation  \cite{med-seg},  volumetric medical image segmentation \cite{volum}, object detection \cite{fcos} and active learning \cite{ac_l}.

Very recently NAS approaches have been proposed for GANs \cite{agan,Autogan}.
These works, which are the closest to ours, use reinforcement learning, which is very different from our approach. 
An important advantage of our approach over them is search time.
The search time in \cite{agan}, which targets directly the inception score that requires class labels information, is extremely high: they use 200 TITAN GPUs for 6 days to perform the search, which is significantly more than the time and computations required by our search. 
The approach in \cite{Autogan}, use RL with parameter sharing and dynamic-resetting. The authors in this work do not perform a search on STL due to search time, which is possible in our proposed strategy due to its better efficiency. 
As we show hereafter, our achieved results are also better than the ones in \cite{agan,Autogan} in terms of FID \cite{fid}. 
\section{GAN search}
\label{Method}
Our goal here is to efficiently search for a suitable generator architecture. As mentioned above, in this work we avoid the search for both a generator and a discriminator due to the instabilities involved in their joint training.  Instead, we make the search simpler by using the GLO method, which does not contain a discriminator. 

Our assumption, which is demonstrated later empirically, is that a generator that is found based on an image reconstruction criterion can also be used to produce improved images when plugged in an existing GAN procedure. 

Our generator search method is inspired by the DARTS algorithm \cite{liu19darts}. It learns in an efficient way what operations to use in each layer of the generator from a pre-fixed set of operations that are all used at the beginning. Before we describe our strategy we briefly survey the DARTS technique and then mention the innovation that allows us to have an efficient search technique for generators. 

\subsection{DARTS}
The DARTS strategy \cite{liu19darts} contains two phases: In the first one, the algorithm searches for the network architecture; and in the second, the new architecture is evaluated after training it from scratch.
The work in \cite{zophNasRL} has noticed that recent convolution neural networks are built from a repeating structure of operations that when concatenated together form a network. 
DARTS follows this routine and in the search phase, it searches for this repeating structure of operations, which is called a 'cell'.
Each cell is composed of a feed-forward graph of feature maps that are connected between them by a mixture of operations.
This \emph{Mixed Operation} denoted by $\bar{o}^{(i,j)}$ is equal to
\begin{equation}\label{eq:mixedOp}
\bar{o}_{i,j}(x)=\frac{\sum_{o\in\mathcal{O}}exp(\alpha_o^{(i,j)})o(x)}{\sum_{o'\in\mathcal{O}}exp(\alpha_{o'}^{(i,j)})},
\end{equation}
where $o(x)$ is an operation on $x$, $\mathcal{O}$ is the group of all operations in the search space, and $\alpha_o^{i,j}$ is the learned weight for the operation $o$.
The $\alpha$ values in \eqref{eq:mixedOp} are learned by optimizing the loss function w.r.t their values. 

In DARTS, two types of cells are being learned - Normal and Reduction.
Normal cell outputs a feature map of the same size as the input.
The reduction cell outputs a feature map of a smaller size than the input.
The operations $o^{(i,j)}$ can be average/max-pool or types of convolutions with varying kernel sizes and strides. The kind of operations that are used depend on the cell type.
The complete network is composed of a concatenation of several cells.

In the search phase, to make the process faster, the network is smaller than the one that will be trained and evaluated in the second phase.
Because the search space is continuous, the optimization process is much faster than prior NAS works.

At the end of the search, pruning is done to most of the operations, based on their $\alpha$ values. The remaining operations (non-pruned ones) are the final cell structure that was learned.
 
\subsection{The generator search}
In our search strategy, we do not use the cells method.
Since the commonly used generator networks are not as deep as the ones used in classification, we can search for the whole network altogether without using cells.
% Although many generator architectures are also built from several repeating operations, we exploit the relatively short length of the network and the efficiency of the search algorithm to make the search more flexible in finding new structures.

{\bf Search phase.} 
In order to search only for a generator, we employ the GLO approach.
This strategy trains a generator alone without the use of a discriminator by optimizing both with respect to the latent space input ($z$) and the generator weights. As we stated before, the advantage of using GLO is its stability, its elimination of the need for a discriminator and its smaller memory use in the search phase- As we do not need to backpropagate gradients from the discriminator. This allows us to find more complex models in a shorter time.
Also, its monotonic stable loss function helps the search to converge.
%}
%The advantage of using GLO in the generator architecture search is that it is more stable compared to regular GAN training. 
%It also allows us to search directly only the generator, which disentangle its search from the discriminator.
%This leads both to better results, and to a smaller memory use in the search phase (as we do not need to backpropagate gradients from the discriminator), which in turn allows finding more complex models in a shorter time. 

%Also, GANs are difficult to train, due to problems such as mode collapse and sensitivity to hyperparameters.
%There are several ways to use a discriminator in the search process - searching for both generator and discriminator, or search only for a generator and use existing discriminator. Learning from other previous attempts, a search for more than one network is a very sensitive and unstable, time-consuming process. Moreover, the first route has the complexities that exist in both NAS and GAN. 
%Thus, we decided to pursue a more stable path in which we just search for a generator and use a simpler loss function. This is the main motivation behind our selection of the GLO objective.

The method works as follow: For each train image it matches a random noise vector $z$ in the latent space, in order to train the generator to output for each z its corresponding image. Then, a reconstruction loss w.r.t the train image (compared to the generator output) is calculated. The reconstruction loss is a combination of a squared-loss function and a Laplacian pyramid (Lap1loss):
\begin{eqnarray}
\label{eq:GLO_losses}
 \ell_{2}(x,\grave{x}) &=&  \parallel x-\grave{x}\parallel^2_2, \\ \nonumber
 Lap_{1}(x,\grave{x}) &=&  \sum_j 2^{2j}\mid L^j (x) - L^j(\grave{x}) \mid_1,
\end{eqnarray}
where $L^j(x)$ is the $j$th level of the Laplacian pyramid.

Using alternating optimization between the GLO and $\alpha$ values, we perform one step of optimization with respect to the latent space input (Z) and generator weights as in GLO, and one step of optimization with respect to the $\alpha$ values of the MixedOp. The loss function is a combination of the losses in \eqref{eq:GLO_losses} as in GLO.

{\bf Searched network structure.} The searched network is divided into three parts. The first is fixed and inspired by ResNet and contains a linear operation with reshape. The second contains the architecture and operations that we search for. The last part is also fixed and contains bn+relu+conv+tanh, again, similar to ResNet.

% As in DARTS, we use MixedOp to connect between feature maps. Per the searched block in the network, our sets of operations are Normal and Up-sample.
% Normal operations are keeping the size of their input and up-sample operations are increasing the size of their input.
% We also prune connections at the end based on their value, as we explain hereafter. 

As in DARTS, we use MixedOp to connect between feature maps but with different types of operations in each MixedOp. We use two types: (i) Normal operations that keep the size of their input; and (ii) up-sample operations that increase the size of their input.
We also prune connections at the end based on their value, as we explain hereafter. 

\begin{figure}
\centering
\includegraphics[width=\linewidth]{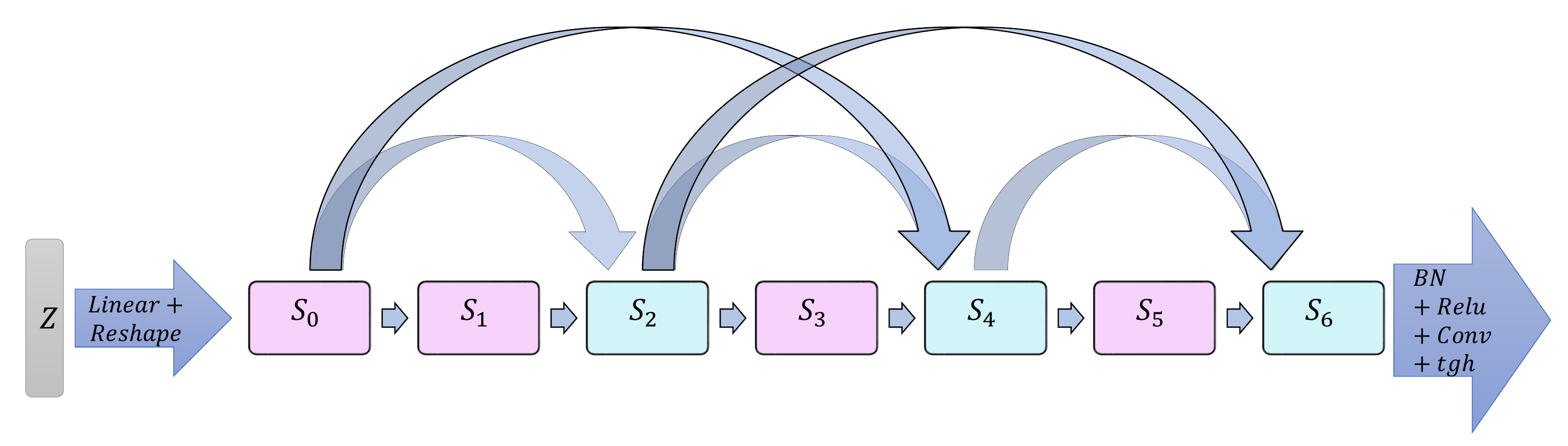}
\caption{The network is built from two types of operations - normal and up-sample. The arrows between the feature maps are learned and called MixedOps. The arrows that go into the blue blocks are up-sample operations and the arrows that go into the pink blocks are normal operations. The architecture of the big arrows is constant }
\label{images/example network}
\end{figure}

The normal operations employed are inspired by DARTS and ResNet architecture. They include a combination of bn+Relu+conv with kernel sizes of 1 and 3, Max and Average pooling, skip connection, separate convolutions(see \cite{liu19darts}) and dilated convolution with kernel sizes of 3 or 5. 

The up-sample operations used are inspired by the DC-GAN and ResNet architectures and include bn+Relu+deconvolution with kernel sizes of 4 and 6 and bn+Relu+nearest neighbor up-sampling + convolution with a kernel size of 1 or 3.
Figure~\ref{images/example network} presents the connections used in the search between the different feature maps. We define normal and up-sampled feature maps as the feature maps that are created using normal and up-sample operations respectively. In other words, the input MixedOp to a normal/up-sampled feature map is a normal/up-sample MixedOp. 
As can be seen in the figure, the up-sampled feature maps are connected between them with residual connections and are connected to the normal feature map before them.
The number of normal feature maps between the up-sample feature maps is determined by the hyperparameter $n$. The figure shows the case of $n$=1, where there is only one normal feature map between two up-sample feature maps. In this case, each normal feature map is connected to the up-sample feature map before it.

% The selection of the up-sample operations are inspired by DC-GAN and resnet architectures, and includes:  combinations of bn+Relu+deconvolution with a kernel size of 4 and 6 and combinations of bn+Relu+nearest neighbor up-sample + convolution with a kernel size of 1 or 3. As can be seen in Figure~\ref{images/example network}, the up-sampled feature maps are connected between them with residual connections and are connected to the previous feature maps. The normal feature maps are connected with a residual-connection to all other normal feature maps from the previous up-sample feature map to them. The number of normal feature maps between the up-sample feature maps is determined by a hyperparameter called $n$. In Figure 1 we present the case of $n=1$.

%%
\begin{figure*}[h!]
        \centering
        \includegraphics[width=0.7\linewidth]{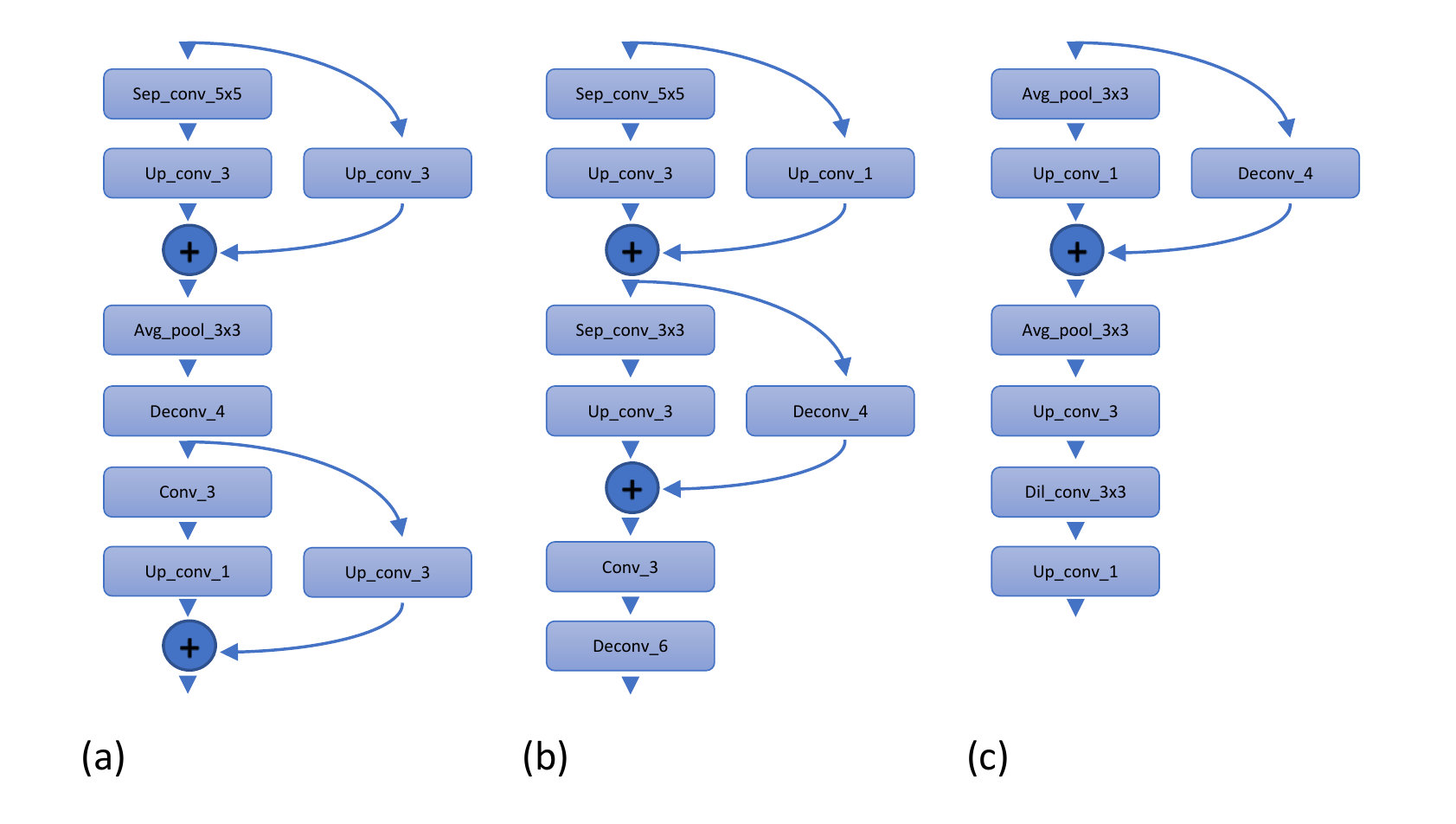}
        \caption{The generators found for: (a) CIFAR-10 ($n$=1); (b) STL ($n$=1); (c) CelebA ($n$=1).}
        \label{fig:generator network cifar10 stl }
\end{figure*}  

\textbf{Pruning phase.} At the end of the search phase, we perform pruning.
The rationale behind our pruning strategy is to allow having skip connections in the network structure but without enforcing them. Because we do not use the same cell configuration as in DARTS (we do not concatenate the outputs), we have to change the pruning procedure. Otherwise, two close nodes may not be connected if we keep using the DARTS pruning procedure. Our new pruning does not enforce any macro architecture but lets the search decide whether to select residual or not, based on the found alpha value. 

The pruning consists of two stages. In the first stage, for each feature map, we keep only one connection to its previous feature maps. The connection is selected to be the operation with the largest value of $\alpha$. In the second stage, if the connection that was selected in the first stage is residual (which may leave the previous feature map unconnected as the operation skips it), we will also add another operation by selecting from the direct connections the one with the highest $\alpha$. 
As shown in Figure~\ref{fig:generator network cifar10 stl }, the search does choose some residuals connections in some connections but not in all of them.

\textbf{Search space complexity.} We turn to analyze the complexity of the search space. 
From looking at Figure~\ref{images/example network}, note that we have $3$ MixedOps with $7$ up-sample operations, $3$ direct MixedOps with $4$ normal operations, and $5$ MixedOps on residual connection with 4+1 up-sample operations each (+1 because they can have lack of connection).
Thus, the number of possible discretized networks (i.e. combination of operations after pruning) is $4^3\cdot7^3\cdot5\cdot9^2$ $\approx$ $10^7$.

The continuous search space (which is the search space before the pruning) has $9^3$ options for normal operations and up-sample operations, $4^3$ options for direct connection and $(4+1)^5$ residual connection (+1 includes the zero operation, that stands for no connection).
Overall, we have $1.458*10^8$ possible networks in this case.

{\bf GAN training.} After finding the generator architecture with our search, we can use it to replace the generator in any given GAN framework, e.g., CTGAN \cite{ctagn}. Then, we simply train the new generator with the discriminator of that framework.

\begin{figure*}  
    \centering
        \includegraphics[width=0.89\linewidth]{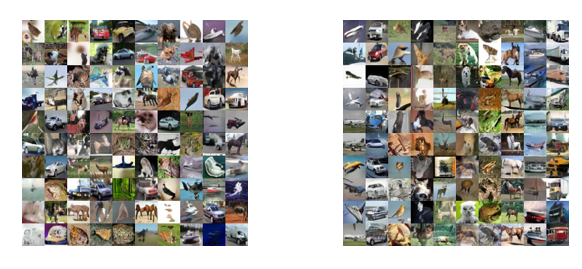}
        \caption{CIFAR-10: (left) unsupervised generated images. (right) supervised generated images}
        \label{fig:cifar10}
    \end{figure*}

\begin{figure*}
    \centering
        \includegraphics[width=0.89\linewidth]{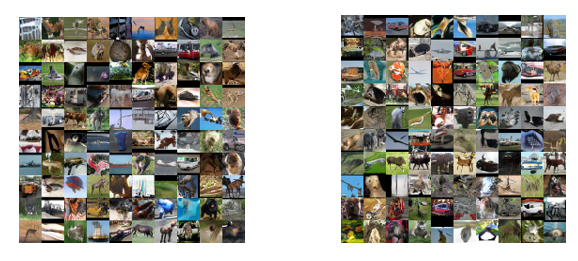}
        \caption{STL unsupervised generated images using: (left) CIFAR-10 based generator (right) STL based generator}
        \label{fig:stl10}
    \end{figure*}
    
\begin{figure*}    
    \centering
        \includegraphics[width=0.89\linewidth]{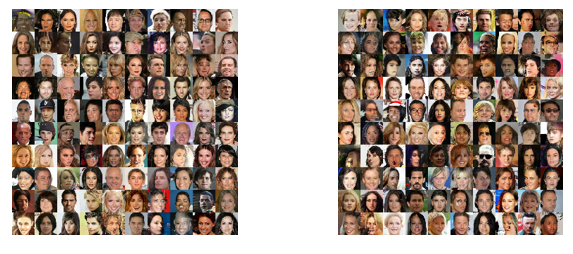}
        \caption{CelebA unsupervised generated images using: (left) CIFAR-10 network (right) CelebA network}
        \label{fig:celebA}
    \end{figure*}

\section{Experiments}
We used CIFAR-10, CelebA and STL datasets for our experiments.
For having a quantitative evaluation, we measured the FID \cite{fid} and Inception score (IS) \cite{IS} on CIFAR-10 and STL.
A larger version of the figures presented here appears in the supp. material.

For calculating the IS, we employed the same method as in \cite{IS_method} and other works and for calculating the FID we used the same method as in \cite{fid}.
In their computation, we generated random 50,000 images with the currently trained model and then use 50,000 real images to calculate the IS and FID.
The inception score is calculated over $10$ splits of the generated images, then mean and std($\pm$ sign) are calculated from these splits.

\subsection{Search results and generalization}
We searched for a generator for CIFAR-10, STL, and CelebA using the scheme described above. In all cases, we do not use the labels provided with the data (in CIFAR-10 and STL) in the search. Yet, we show hereafter that the generator architecture that is found without labels shows good performance also when the labels are added. We show also the transferability of the model across datasets. 

We use the DARTS \cite{liu19darts} hyperparameters of CIFAR-10, except for the learning rate that is set to 3e-1 for CIFAR-10, and 3e-2 for CelebA and STL.

On 1 Nvidia TITAN-X, the search took 28 hours for CIFAR-10, 100 hours for STL, and 57 hours for CelebA.
The unsupervised generator training time was 37 hours for CIFAR-10, 85 hours for STL and 33 hours for CelebA. 

{\bf Models size.}
The model size of the CIFAR10 generator is 1.65 MB and the STL generator is 1.17 MB.

{\bf CIFAR-10.} Figure~\ref{fig:generator network cifar10 stl } presents the network found for CIFAR-10. Notice that different operations are selected in each layer (both for the up-sample and normal operations), which demonstrates the advantage of not forcing the network into the cell structure. 
Note also that the found architecture does contains skip connections, which resemble the current state-of-the-art generators.
Table~\ref{unsupervised cifar10} compares the IS and FID scores of our model to other existing works. Notice that our results are on par with the state of the art methods and better than the other searched GANs (we are on the same scale like \cite{agan} in terms of IS but much better than it in terms of FID, which is considered to be a better perceptual measure \cite{fid}. Thus, we may claim that our result is better).

{\bf Supervised CIFAR-10.} For CIFAR-10, we also train the found architecture in a supervised way, following the procedure in \cite{ctagn}.
By that, we demonstrate that the model we have found is transferable to supervised training.
The results of the supervised case appear in Table~\ref{supervised cifar10}.
Note that although the generator was searched in an unsupervised form, our supervised results are competitive with the other supervised generators and better as before than AGAN \cite{agan} (no result was reported for this case for AutoGAN in \cite{Autogan}).
% The supervised train takes 41 hours for a single network 

\begin{table}
    \begin{center}
        \begin{tabular}{l|c|c|c}
        % \hline
            Method 
            & \multicolumn{1}{|p{1.45cm}}{\centering Inception Score }
            & \multicolumn{1}{|p{1.2cm}}{\centering FID}
            & \multicolumn{1}{|p{1.2cm}}{\centering \small{Search} \\ \small{GPU days} }
            \\
            \hline
            Real data 
            & 11.24 
            & 2.1 
            & - \\
            \hline
            SN-GAN \cite{sngan}  & 8.22 $\pm$ 0.5 & 11.8* &  \\
            WGAN-GP \cite{wgan-gp} & 7.86 $\pm$ 0.07 & 14.1* & - \\
            CTGAN \cite{ctagn}  & 8.12 $\pm$ 0.12  & - & - \\ 
            CAGAN \cite{cagan}  & 8.35 $\pm$ 0.09  & - & - \\
            AGAN \cite{agan}  & 8.29 $\pm$ 0.9 & 30.5 & 1200 \\                 
            AutoGAN \cite{Autogan}  & 8.55 $\pm$ 0.1 & 12.42 & n/a \\  
            \hline
            \ourmethod{}(ours)
            & 8.37 $\pm$ 0.08
            & 12.01
            & 1.167 
            \end{tabular}
    \end{center}
    \caption{CIFAR-10 unsupervised image generation. *FID taken from \cite{howgood}}
    \label{unsupervised cifar10}
\end{table}

\begin{table}
    \begin{center}
        \begin{tabular}{l|c|c|c}
            Method 
            & \multicolumn{1}{|p{1.45cm}}{\centering Inception Score }
            & \multicolumn{1}{|p{1.2cm}}{\centering FID}
            & \multicolumn{1}{|p{1.2cm}}{\centering Search \\ GPU days }
            \\
            \hline
            Real data 
            & 11.24 
            & 2.1 
            & - \\
            \hline
            WGAN-GP \cite{wgan-gp} & 8.42 $\pm$ 0.1 & - & - \\
            CAGAN \cite{cagan}  & 8.89 $\pm$ 0.11   & - & - \\ 
            CTGAN \cite{ctagn}  & 8.81 $\pm$ 0.13   & - & - \\ 
            AGAN \cite{agan}  & 8.82 $\pm$ 0.9 & 23.8 & 1200 \\                 
            
            \hline
            \ourmethod{}(ours) 
            & 8.85 $\pm$ 0.07
            & 9.83
            & 1.167
            \end{tabular}
    \end{center}
    \caption{CIFAR-10 supervised image generation}
    \label{supervised cifar10}
\end{table}

\begin{table*}
    \begin{center}
        \begin{tabular}{l|c|c|c}
            Method 
            & \multicolumn{1}{|p{1.45cm}}{\centering Inception Score }
            & \multicolumn{1}{|p{1.45cm}}{\centering FID}
            & \multicolumn{1}{|p{1.45cm}}{\centering Search cost \\ GPU days }
            \\
            \hline
            Real data 
            & 26.08 $\pm$ 0.26
            & 3.5
            & - \\
            \hline
            SN-GAN \cite{sngan}  & 9.10 $\pm$ 0.04  & 40.1 & - \\
            WGAN-GP \cite{wgan-gp} & 9.05 $\pm$ 0.12 & - & - \\
            CAGAN \cite{cagan}  & 9.51 $\pm$ 0.14 & - & - \\ 
            AGAN \cite{agan}  & 9.23 $\pm$ 0.08 & 52.7 & 1200 \\    
            AutoGAN \cite{Autogan}  & 9.16 $\pm$ 0.12 & 31.01 & n/a \\      
            \hline
            \ourmethod{}(ours) - searched & 9.22 $\pm$ 0.08& 40.25 & 4.167 \\
            \ourmethod{}(ours) - CIFAR-10 net & 9.71 $\pm$ 0.11 & 28.76& 1.167 
            \end{tabular}
    \end{center} 
    \caption{STL unsupervised image generation. The first row indicates the score for a network searched on STL dataset. The second row indicates the score for a network searched on CIFAR-10.}
    \label{unsupervised stl10}
\end{table*}

{\bf STL.} For STL, we performed two experiments: (i) searching on STL and then training on it; and (ii) taking the network found on CIFAR-10 and training it on STL. As the image size in CIFAR-10 and STL is different, we have increased the latent vector size by a factor of the image size ratio between CIFAR-10 and STL,
in the CIFAR-10 network, which leads to the desired size at the output for STL.

As Table~\ref{unsupervised stl10} shows, the architecture we found on CIFAR-10 transfers well to STL and even outperforms the one searched on STL.
We believe that the difference between the results is due to the hyperparameters (initially designed for CIFAR-10) used in the search. We expect that a better hyperparameter setting will improve the STL search. Note that our FID score is better than the compared methods. 

Note that the fact that we can search on STL makes us conclude that our search time is better than the one of AutoGAN. Although its search time is not reported, it was claimed that they did not search on STL due to the search time. Note that our search time on STL is just a few days on a single GPU, which is a reasonable time.

{\bf CelebA.} Similar to STL, also for CelebA we considered two generators: One that was searched on CelebA images of size 32x32; and one that was searched on CIFAR-10. Figure~\ref{fig:celebA} presents the 32x32 generated images. Notice that the quality of images is similar for both generators, which shows the transferability of the generator found on CIFAR-10 to a different type of images.

\subsection{Ablation study}
\begin{table}
    \begin{center}
        \begin{tabular}{l|c}
            Method 
            & \multicolumn{1}{|p{1.45cm}}{\centering Inception Score }
            \\
            \hline
            Search with $n$ = 2 & 7.24 $\pm$ 0.06 \\
            Images size 16x16 & 8.03 $\pm$ 0.012\\
            Random block  &  8.127 $\pm$ 0.098\\ 
            \hline
            \ourmethod{}(ours) &  8.37 $\pm$ 0.08\\ 
            \end{tabular}
    \end{center}
    \caption{Ablation studies on CIFAR-10}
    \label{ablation}
\end{table}

{\bf Training random block.}
Taking random continues architecture, pruning it and then applying it to CIFAR-10 leads to IS and FID of 8.127 $\pm$ 0.098, 15.312, which is significantly inferior to our found network. This further confirms the effectiveness of \ourmethod{}.

{\bf Effect of different $n$ values.}
We evaluate the effect of the number of normal feature maps ($n$) between up-sample feature maps on the search.
When searching on CIFAR-10 using $n$=2, the IS is low (7.24 $\pm$ 0.06) and the FID is high (48.83), which shows that $n$=1 is better suited for the generator task. The found generator can be found in the supp. material.

{\bf Effect of search with smaller image size.} We also check whether decreasing the size of the images in the search affects the performance of the found architecture. 
For CIFAR-10, we searched using images with a size 16x16 and trained on regular-sized images. On STL we searched using images of size 32x32 and then trained the found generator on regular images (of size 48x48). Although the search time was smaller, the IS was much lower for both (8.03 $\pm$ 0.012 and 8.97 $\pm$ 0.02).

\section{Conclusion}
\label{Discussion and conclusions}
This paper introduced \ourmethod{}, a method for searching efficiently generator architectures without the need to search for a discriminator.
On CIFAR-10 and STL, \ourmethod{} outperforms parallel works to us for automated GAN search \cite{agan},\cite{Autogan} both in terms of the search cost and terms of FID score.
We have also demonstrated the transferability of our found generator both across datasets and between unsupervised and supervised training. 

The paper's main contribution is providing an efficient generator search strategy.
It avoids long search time by using a  continuous search space and the GLO framework.
% As it is the first work in this direction, 
We believe our work has various interesting follow-up GAN-related search directions. We mention two of them: 

1. Most GANs generate low-resolution images. A solution to this problem is to train GAN progressively \cite{prog_gan}. This means that we start with a generator and discriminator for small image output, and after stabilizing the network, we progressively expand the network output.
A possible future work is combining our search with the progressive GAN framework to create a search strategy for high-resolution GANs and further improve results.
%We didn't use progressive GAN methods for our search, but we think it can gain further improvement in results. 

2. In this work, we did not select the discriminator, which may have a great impact on the results.
Future work should explore also this important aspect of GAN training. Discriminators are mostly built from down-sample and normal operations as in the DARTS searched classifiers and can be added as another optimizing phase to our scheme, before or together with the generator search.

{\bf Acknowledgment.} This work was supported by Alibaba and the NSF-BSF grant.

{\small
\bibliographystyle{ieee_fullname}
\bibliography{my_bib}
}
\end{document}